\documentclass{article} 
\usepackage{iclr2016_conference,times}
\usepackage{hyperref}
\usepackage{url}
\usepackage{graphicx}
\usepackage{amsmath}
\usepackage{multirow}
\usepackage{graphicx}
\usepackage{float}
\usepackage[flushleft]{threeparttable}

\title{AUC-maximized Deep Convolutional Neural Fields for Sequence Labeling}

\author{
Sheng Wang$^{1,2}$, Siqi Sun$^1$ \& Jinbo Xu$^1$ \\
1. Toyota Technological Institute at Chicago, Chicago, IL\\
2. Department of Human Genetics, University of Chicago, Chicago, IL\\
\texttt{wangsheng@uchicago.edu, \{siqi.sun, j3xu\}@ttic.edu} \\
}

\begin{document}
\maketitle

\begin{abstract}
Deep Convolutional Neural Networks (DCNN) has shown excellent performance in a variety of machine learning tasks. This paper presents Deep Convolutional Neural Fields (DeepCNF), an integration of DCNN with Conditional Random Field (CRF), for sequence labeling with an imbalanced label distribution. 
The widely-used training methods, such as maximum-likelihood and maximum labelwise accuracy, do not work well on imbalanced data. 
To handle this, we present a new training algorithm called maximum-AUC for DeepCNF. 
That is, we train DeepCNF by directly maximizing the empirical Area Under the ROC Curve (AUC), which is an unbiased measurement for imbalanced data. To fulfill this, we formulate AUC in a pairwise ranking framework, approximate it by a polynomial function and then apply a gradient-based procedure to optimize it. We then test our AUC-maximized DeepCNF on three very different protein sequence labeling tasks: solvent accessibility prediction, 8-state secondary structure prediction, and disorder prediction. Our experimental results confirm that maximum-AUC greatly outperforms the other two training methods on 8-state secondary structure prediction and disorder prediction since their label distributions are highly imbalanced and also has similar performance as the other two training methods on solvent accessibility prediction, which has three equally-distributed labels. Furthermore, our experimental results show that our AUC-trained DeepCNF models greatly outperform existing popular predictors of these three tasks.
\end{abstract}

\section{Introduction}
Deep Convolutional Neural Networks (DCNN), originated by Yann LeCun at 1998 ~\citep{lecun1998gradient} for document recognition, is being widely used in a plethora of machine learning (ML) tasks ranging from speech recognition ~\citep{hinton2012deep}, to computer vision~\citep{krizhevsky2012imagenet}, and to computational biology~\citep{di2012deep}. DCNN is good at capturing medium- and/or long-range structured information in a hierarchical manner. To handle structured data, \citet{chen2014semantic} has integrated DCNN with fully connected Conditional Random Fields (CRF) for semantic image segmentation. Here we present Deep Convolutional Neural Fields (DeepCNF), which is an integration of DCNN and linear-chain CRF, to address the task of sequence labeling and apply it to three important biology problems: solvent accessibility prediction (ACC), disorder prediction (DISO), and 8-state secondary structure prediction (SS8)~\citep{magnan2014sspro,jones2015disopred3}. See Appendix for a brief description of these problems.

A protein sequence can be viewed as a string of amino acids (also called residues in the protein context) and we want to predict a label for each residue. In this paper we consider three types of labels: solvent accessibility, disorder state and 8-state secondary structure. These three structure properties are very important to the understanding of protein structure and function. The solvent accessibility is important for protein folding~\citep{dill1990dominant}, the order/disorder state plays an important role in many biological processes~\citep{oldfield2014intrinsically}, and protein secondary structure(SS) relates to local backbone conformation of a protein sequence~\citep{pauling1951structure}. 
The label distribution in these problems varies from almost uniform to highly imbalanced. For example, only $\sim$6\% of residues are shown to be disordered~\citep{he2009predicting}. Some SS labels, such as 3-10 helix, beta-bridge, and pi-helix are extremely rare~\citep{wang2011protein}. The widely-used training methods, such as maximum-likelihood~\citep{lafferty2001conditional} and maximum labelwise accuracy~\citep{gross2006training}, perform well on data with balanced labels but not on highly-imbalanced data~\citep{de2012weighted}.

This paper presents a new maximum-AUC method to train DeepCNF for imbalanced sequence data. Specifically, we train DeepCNF by maximizing Area Under the ROC Curve (AUC), which is a good measure for class-imbalanced data~\citep{cortes2004auc}. Taking disorder prediction as an example, random guess can obtain $\sim$94\% per-residue accuracy, but its AUC is only $\sim$0.5. AUC is insensitive to changes in class distribution because the ROC curve specifies the relationship between false positive (FP) rate and true positive (TP) rate, which are independent of class distribution~\citep{cortes2004auc}. However, it is very challenging to directly optimize AUC. A few algorithms have been developed to maximize AUC on unstructured data~\citep{joachims2005support,herschtal2004optimising,narasimhan2013structural}, but to the best of our knowledge, there is no such an algorithm for imbalanced structured data (e.g., sequence data addressed here). To train DeepCNF by maximum-AUC, we formulate the AUC function in a ranking framework, approximate it by a polynomial Chebyshev function~\citep{calders2007efficient} and then use L-BFGS~\citep{liu1989limited} to optimize it.

Our experimental results show that when the label distribution is almost uniform, there is no big difference between the three training methods. Otherwise, maximum-AUC results in better AUC and Mcc than the other two methods. Tested on several publicly available benchmark data, our AUC-trained DeepCNF model obtains the best performance on all the three protein sequence labeling tasks. In particular, at a similar specificity level, our method obtains better precision and sensitivity for those labels with a much smaller occurring frequency.

\textbf{Contributions.}
1. A novel training algorithm that directly maximizes the empirical AUC to learn DeepCNF model from imbalanced structured data.
2. Studying three training methods, i.e. maximum-likelihood, maximum labelwise accuracy, and maximum-AUC, for DeepCNF and testing them on three real-world protein sequence labeling problems, in which the label distribution varies from almost uniform to highly imbalanced.
3. Achieving the state-of-the-art performance on three important protein sequence labeling problems.

\section{Related work}
The maximum-AUC training method is not a totally new idea. There are already some studies on un-structured data, e.g., (a) \citet{ferri2002learning} trained a decision tree using AUC as a splitting criterion; (b)\citet{herschtal2004optimising} trained a neural network by optimizing AUC; (c)\citet{joachims2005support} described a generalized Support Vector Machines (SVM) that optimizes AUC; and (d)\citet{narasimhan2013structural} explored ways to optimize partial AUC for a structured SVM. However, we would like to develop a maximum-AUC training method for DeepCNF to handle sequential data.

Recently, Rosenfeld et al  has presented a AUC-based learning algorithm for structured models\citet{rosenfeld2014learning}, targeting at a ranking problem. Our method differs from this work as follows: (a) our method targets at a sequence labeling problem with imbalanced label distribution, but not a ranking problem; (b) we consider correlation among labels in a sequence while Rosenfeld et al treat the ranking of each sample independent of the others; (c) we work on DeepCNF while they use structured SVM.

\section{Method}
\subsection{DeepCNF model}

\begin{figure}[h]
\begin{center}
\includegraphics[width=100mm]{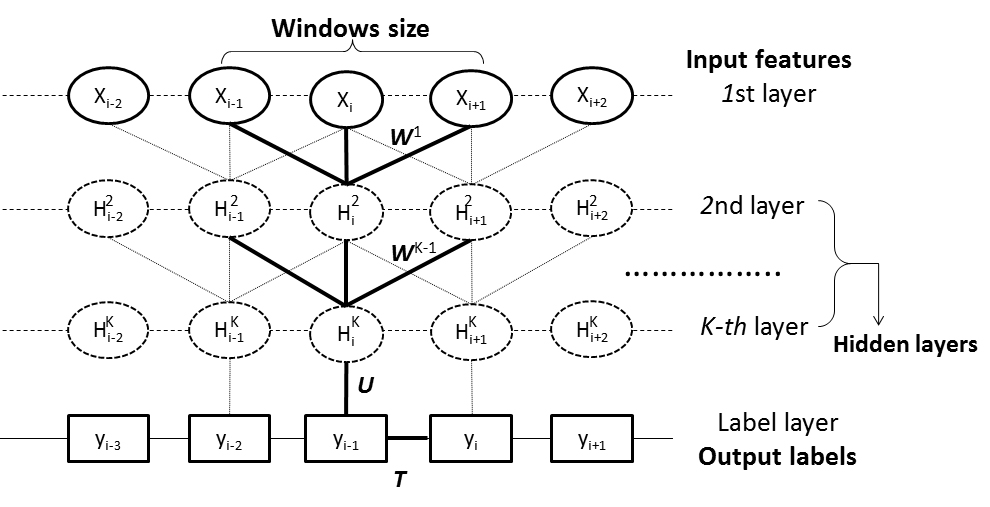}
\end{center}
\caption{Illustration of a DeepCNF. Here $i$ is the position index and $X_i$ the associated input features, $H^k$ represents the $k$-th hidden layer, and $Y$ is the output label. All the layers from the first to the top layer form a DCNN with parameter $W^k\{k=1, \dots, K\}$. The top layer and the label layer form a CRF, in which the parameter $U$ specifies the relationship between the output of the top layer and the label layer and $T$ is the parameter for adjacent label correlation. Windows size is set to 3 only for illustration.}
\label{fig:DeepCNF}
\end{figure}

As shown in Figure \ref{fig:DeepCNF}, DeepCNF has two modules: (i) the Conditional Random Fields (CRF) module consisting of the top layer and the label layer, and (ii) the deep convolutional neural network (DCNN) module covering the input to the top layer. When only one hidden layer is used, DeepCNF becomes Conditional Neural Fields (CNF), a probabilistic graphical model described in \citet{peng2009conditional}.

Given a sequence of length $L$, let $y=(y_1, \dots, y_L) \in \Sigma^L$ denote its sequence label where $y_i$ is the label at residue $i$, and $\Sigma$ is the set of all possible labels. For instance, for protein disorder prediction, $\Sigma = \{0, 1\}$ where $0$ stands for ordered and $1$ for disordered. Let $X = (X_1, X_2, \dots, X_L)$ denote the input feature where $X_i$ is a column vector representing the input feature for position $i$. DeepCNF calculates the conditional probability of $y$ on the input $X$ with parameter $\theta$ as follows,
\begin{align}
P_\theta (y|X) = \frac{1}{Z(X)} \exp \Big( \sum_{i=1}^L ( f_\theta(y,X,i) + g_\theta(y, X, i)) \Big),
\label{eq:model}
\end{align}
where $f_\theta(y, X, i)$ is the binary potential function specifying correlation among adjacent labels at position $i$, $g_\theta(y, X, i)$ is the unary potential function modeling relationship between $y_i$ and input features for position $i$, and $Z(X)$ is the partition function. Formally, $f_\theta(\cdot)$ and $g_\theta(\cdot)$ are defined as follows:
\begin{alignat*}{2}
f_\theta(y, X, i) =& f_\theta(y_{i-1}, y_i, X, i)  &=& \sum_{a,b} T_{a,b} \delta(y_{i-1}=a) \delta(y_i=b)， \\
g_\theta(y, X, i) =& g_\theta(y_i, X, i)  &=& \sum_{a,h} U_{a,h} A_{a,h}(X,i,W) \delta(y_i=a),
\end{alignat*}
where $a$ and $b$ represent two specific labels for prediction, $\delta(\cdot)$ is an indicator function, $A_{a,h}(X,i,W)$ is a deep neural network function for the $h$-th neuron at position $i$ of the top layer for label $a$, and $W, U$ and $T$ are the model parameters to be trained. Specifically, $W$ is the parameter for the neural network, $U$ is the parameter connecting the top layer to the label layer, and $T$ is for label correlation. The two potential functions can be merged into a single binary potential function $f_\theta(y,X,i) = f_\theta(y_{i-1},y_i,X,i) = \sum_{a,b,h} T_{a,b,h} A_{a,b,h}(X,i,W)\delta(y_{i-1}=a) \delta(y_i=b)$. Note that these deep neural network functions for different labels could be shared to $A_h(X, i, W)$. To control model complexity and avoid over-fitting, we add a $L_2$-norm penalty term as the regularization factor. 

Figure \ref{fig:DeepCNF} shows two adjacent layers of DCNN. Let $M_k$ be the number of neurons for a single position at the $k$-th layer. Let $X_i(h)$ be the $h$-th feature at the input layer for residue $i$ and $H_i^k(h)$ denote the output value of the $h$-th neuron of position $i$ at layer $k$. When $k=1$, $H^k$ is actually the input feature $X$. Otherwise, $H^k$ is a matrix of dimension $L\times M_k$. Let $2N_k+1$ be the window size at the $k$-th layer. Mathematically, $H_i^k(h)$ is defined as follows:
\begin{align*}
H_i^k(h) =& X_i(h), && \text{if } k = 1 \\
H_i^{k+1}(h) =& \pi\Big( \sum_{n=-N_k}^{N_k} \sum_{h=1}^{M_k} ( H_{i+n}^k(h) * W_n^k(h,h') ) \Big)  &&\text{if } k < K\\
A_h(X, i, W) = &H_i^k(h)   &&\text{if } k = K.
\end{align*}
Meanwhile, $\pi(\cdot)$ is the activation function, either the sigmoid (i.e. $1/(1+\exp(-x))$) or the tanh (i.e. $(1-\exp(-2x))/(1+\exp(-2x))$) function. $W_n^k (-N_k \le n \le N_k)$ is a 2D weight matrix for the connections between the neurons of position $i+n$ at layer $k$ and the neurons of position $i$ at layer $k+1$. $W_n^k(h, h')$ is shared by all the positions in the same layer, so it is position-independent. Here $h'$ and $h$ index two neurons at the $k$-th and $(k+1)$-th layers, respectively.  See Appendix about how to calculate the gradient of DCNN by back propagation.

\section{Training Methods} 
Let $T$ be the number of training sequences and $L_t$ denote the length of sequence $t$. 
We study three different training methods: maximum-likelihood, maximum labelwise accuracy, and maximum-AUC.

\subsection{Maximum-likelihood}
The log-likelihood is a widely-used objective function for training CRF~\citep{lafferty2001conditional}. Mathematically, the log-likelihood is defined as follows:
\begin{align*}
LL = \sum_{t=1}^T \log P_\theta (y^t | X^t),
\end{align*}
where $P_\theta(y|X)$ is defined in equation (\ref{eq:model}). 

\subsection{maximum labelwise accuracy}
\citet{gross2006training} proposed an objective function that could directly maximize the labelwise accuracy defined as
\begin{align*}
LabelwiseAccuracy = \sum_{t=1}^T \sum_{i=1}^{L_t} \delta\Big( P_\theta(y_i^{(\tau)}) > \max_{y_i \ne y_i} P_\theta(y_i) \Big),
\end{align*}
where $y_i^{(\tau)}$ denotes the real label at position $i$, $P_\theta(y_i^{(\tau)})$ is the predicted probability of the real label at position $i$. It could be represented by the marginal probability 
\begin{align*}
P_\theta( y_i^{(\tau)} | X^t) = \frac{1}{Z(X)} \sum_{y_{1:L^t}} \big( \delta(y_i=(\tau)) \exp(F_{1:L^t} (y, X^t, \theta) ) \big),
\end{align*}
where $F_{l_1:l_2}(y,X,\theta) = \sum_{i=l_1}^{l_2} f_\theta(y, X, i)$. 

To obtain a smooth approximation to this objective function, \citet{gross2006training} replaces the indicator function with a sigmoid function  $Q_\lambda(x) = 1/(1+\exp(-\lambda x))$ where the parameter $\lambda$ is set to 15 by default. Then it becomes the following form:
\begin{align*}
LabelwiseAccuracy \approx \sum_{t=1}^T \sum_{i=1}^{L_t} Q_\lambda \big( P_\theta( y_i^{(\tau)} | X^t) - P_\theta( \tilde{y}_i^{(\tau)} |X^t) \big),
\end{align*}
where $\tilde{y}_i^{(\tau)}$ denote the label other than $y_i^{(\tau)}$ that has the maximum posterior probability at position $i$. 

\subsection{Maximum-AUC}
The AUC of a predictor function $P_\theta$ on label $\tau$ is defined as:
\begin{align}
AUC(P_\theta, \tau) = P\Big( P_\theta(y_i^\tau) > P_\theta(y_j^\tau) | i\in D^\tau, j\in D^{!\tau} \Big),
\label{eq:auc}
\end{align}
where $P(\cdot)$ is the probability over all pairs of positive and negative examples, $D^\tau$ is a set of positive examples with true label $\tau$, and $D^{!\tau}$ is a set of negative examples with true label not being $\tau$. Note that the union of $D^\tau$ and $D^{!\tau}$ contains all the training sequence positions, i.e., $D^\tau =\cup_{t=1}^T \cup_{i=1}^{L_t} \delta_{i,t}^\tau $ where $\delta_{i,t}^\tau$ is an indicator function. If the true label of the $i$-th position from sequence $t$ equals to $\tau$, then $\delta_{i,t}^\tau$ is equal to 1; otherwise 0. Again, $P_\theta(y_i^\tau)$ could be represented by the marginal probability $P_\theta( y_i^\tau | X^t)$ from the training sequence $t$. Since it is hard to calculate the derivatives of equation (\ref{eq:auc}), we use the following Wilcoxon-Mann-Whitney statistic~\citep{hanley1982meaning}, which is an unbiased estimator of $AUC(P_\theta, \tau)$:
\begin{align}
AUC^{WMW}(P_\theta, \tau) = \frac{ \sum_{i\in D^\tau} \sum_{j\in D^{!\tau}} \delta( P_\theta(y_i^\tau | X) ) > P_\theta(y_j^\tau) | X) }{|D^\tau| | D^{!\tau} |}.
\label{eq:auc-wmw}
\end{align}
Finally, by summing over all labels, the overall AUC objective function is $\sum_\tau AUC^{WMW}(P_\theta, \tau)$. 

For a large dataset, the computational cost of AUC by equation (\ref{eq:auc-wmw}) is high. Recently, \citet{calders2007efficient} proposed a polynomial approximation of AUC which can be computed in linear time. The key idea is to approximate the indicator function $\delta(x>0)$, where $x$ represents $P_\theta(y_i^\tau | X) - P_\theta(y_j^\tau|X)$ by a polynomial Chebyshev approximation. That is, we approximate $\delta(x>0)$ by $\sum_{\mu = 0}^d c_\mu x^\mu$ where $d$ is the degree and $c_\mu$ the coefficient of the polynomial~\citep{calders2007efficient}.  Let $n_1 = |D^\tau|$ and $n_0=|D^{!\tau}|$. Using the polynomial Chebyshev approximation, we can approximate equation (\ref{eq:auc-wmw}) as follows:
\begin{align*}
AUC^{WMW}(P_\theta, \tau) \approx \frac{1}{n_0 n_1} \sum_{\mu=0}^d \sum_{l=0}^\mu  \mathcal{Y}_{\mu l} s(P_\theta^l, D^\tau) v(P_\theta^{\mu-l}, D^{!\tau})
\end{align*}
where $\mathcal{Y}_{\mu l} = c_\mu\binom{\mu}{l} (-1)^{\mu-l}$, $s(P^l, D^\tau) = \sum_{i\in D^\tau} P(y_i^\tau)^l$ and $v(P^l, D^{!\tau}) = \sum_{j\in D^{!\tau}} P(y_j^\tau)^l$. Note that we have $s(P^l, D^\tau) = \sum_{t=1}^T \sum_{i=1}^{L_t} \delta_{i,t}^\tau P(y_i^\tau)^l$ and a similar structure for $v(P^l, D^{!\tau})$.

\subsection{Gradient of the polynomial approximation of AUC}

The gradient of the approximate AUC with respect to the parameter $\theta$ is as follows:
\begin{align*}
\frac{ \partial AUC^{WMW}(P_\theta, \tau) } { \partial \theta} = \frac{1}{n_0 n_1} \sum_{\mu=0}^d \sum_{l=0}^\mu \mathcal{Y}_{\mu l} \Big( \frac{ \partial s(P_\theta^l, D^\tau)}{\partial \theta}v(P_\theta^{\mu-l},D^{!\tau} + s(P_\theta^l, D^\tau) \frac{v(P_\theta^{\mu-l}, D^{!\tau})}{\partial \theta } \Big). 
\end{align*}
Note that the calculation of $\frac{ \partial s(P_\theta^l, D^\tau)}{\partial \theta}$ and $\frac{v(P_\theta^{\mu-l}, D^{!\tau})}{\partial \theta }$ is similar, so we only explain one of them, and suppose there is only one training sequence with length $L$.  In particular,
\begin{align*}
\frac{ \partial s(P_\theta^l, D^\tau)}{\partial \theta} = \sum_{i=1}^L \frac{ \partial (\delta_i^\tau P_\theta(y_i^\tau | X))^l}{\partial \theta}.
\end{align*}
Let $Q_i(P_\theta) = (\delta_i^\tau P_\theta(y_i^\tau | X))^l$, then 
\begin{align}
\frac{ \partial s(P_\theta^l, D^\tau)}{\partial \theta} = \sum_{i=1}^L Q_i' \frac{ \partial P_\theta(y_i^\tau | X)}{\partial \theta},
\label{eq:grad_rule}
\end{align}
where $Q_i'$ is the gradient of $Q_i$ with respect to the marginal probability $P_\theta$.

Since 
\[
P_\theta(y_i^\tau|X) = \frac{1}{Z(X)} \sum_{y_{1:L}} \Big( \delta(y_i=\tau) \exp(F_{1:L} (y, X, \theta)) \Big), 
\]
applying the quotient rule we can compute the gradient of equation (\ref{eq:grad_rule}) as follows
\begin{align}
\frac{ \partial s(P_\theta^l, D^\tau)}{\partial \theta} = \sum_{i=1}^L & \Bigg( \frac{1}{Z(x)} Q_i' \sum_{y_{1:L}} \Big( \delta(y_i=\tau) \frac{\partial F_{1:L}(y,X,\theta)}{\partial \theta}\exp(F_{1:L}(y,X,\theta)) \Big)\Bigg) \nonumber \\
\cdot &\frac{-1}{Z(X)} \frac{\partial Z(x)}{\partial \theta} \sum_{i=1}^L \Big(Q_i' P_\theta(y_i^\tau | X) \Big).
\label{eq:grad}
\end{align}
The second term in equation (\ref{eq:grad}) could be calculated efficiently using forward-backward algorithm. For parameter $T$ at position $i$, the gradient could be calculated as follow:
\begin{align*}
- \mathcal{C} \sum_{u'}\sum_{u} \frac{\alpha(u', i-1)\beta(u,i)}{Z(X)} \exp(f_\theta(u', u, X, i))  \frac{\partial f_\theta(u', u, X, i)}{\partial \theta}.
\end{align*}
For parameter $U$ at position $i$, the gradient could be calculated as follows:
\begin{align*}
-\mathcal{C} \sum_u \frac{\alpha(u, i)\beta(u,i)}{Z(X)} \frac{\partial g_\theta(u, X, i)}{\partial \theta},
\end{align*}
where $u = \sum_{i=1}^L \Big( Q_i' P_\theta(y_i^\tau | X) \Big)$ denotes one label and 
\begin{align*}
\mathcal{C} = \sum_{i=1}^L \Big( Q_i' P_\theta(y_i^\tau |X) \Big).
\end{align*}

The forward function $\alpha(u,i)$ and backward function $\beta(u,i)$ are defined as
\begin{align*}
\alpha(u,i) =& \sum_{y_{1:i}} \delta(y_i=u)\exp( F_{1:i}(y,X,\theta) ) \\
\beta(u,i) =& \sum_{y_{i:L}} \delta(y_i=u)\exp(F_{i+1:L}(y,X,\theta) ).
\end{align*}
They can be calculated by dynamic programming as follows,
\begin{align*}
\alpha(u,i) =& \sum_{u'} \alpha(u', i-1)\exp( f_\theta(u',u, X, i) ) \\
\beta(u, i) =& \sum_{u'} \beta(u', i+1)\exp( f_\theta(u, u', X, i+1)).
\end{align*}
The gradient of the inner summation part of the first term in equation (\ref{eq:grad}) with respect to parameter $T$ at position $i$ could be calculated as follows:
\begin{align*}
\sum_u \sum_{u'} \phi(u', u, i) \exp(f_\theta(u', u, X, i) ) \frac{\partial f_\theta(u', u, X, i) }{\partial \theta},
\end{align*}
where
\begin{align*}
\phi(u',u,i) = Q_i' \delta(y_i = \tau) \frac{\alpha(u', i-1) \beta(u,i)}{Z(X)} + \frac{\alpha^\tau(u', i-1) \beta(u,i)}{Z(X)} + \frac{\alpha(u', i-1) \beta^\tau(u,i)}{Z(X)}.
\end{align*}
Similarly, the inner summation part of the first term in equation (\ref{eq:grad}) with respect to parameter $U$ at position $i$ could be calculated as 
\[
\sum_u \Phi(u,i) \frac{\partial g_\theta(u, X, i)}{\partial \theta},
\]
where $\Phi(u,i) = \frac{\alpha^\tau(u,i)\beta(u,i)}{Z(X)} + \frac{\alpha(u,i)\beta^\tau(u, i)}{Z(X)}$. Here we define,
\begin{align*}
\alpha^\tau(u, i) = & \sum_{t=1}^i \sum_{y_{1:i}} \delta(y_t=\tau \wedge y_i=u) Q_t' \exp(F_{1:i}(y,X,\theta) ) \\
\beta^\tau(u, i) = & \sum_{t=i+1}^L \sum_{y_{i:L}} \delta(y_t=\tau \wedge y_i=u) Q_t' \exp(F_{i+1:L}(y,X,\theta) ).
\end{align*}
Like the forward matrix $\alpha(u,i)$ and backward matrix $\beta(u,i)$, $\alpha^\tau(u,i)$ and $\beta^\tau(u,i)$ may also be calculated by dynamic programming. In particular, given the initial conditions $\alpha^\tau(u,1) = Q_1'\delta(u=\tau) \alpha(u,1)$ and $\beta^\tau(u,L) = 0$. $\alpha^\tau(u,i)$ and $\beta^\tau(u,i)$ can be computed by the following recurrences:
\begin{align*}
\alpha^\tau(u, i) =& \sum_{u'} \Big( \alpha^\tau(u', i-1) + Q_i'\delta(u=\tau)\alpha(u',i-1)\Big)\exp( f_\theta(u',u,X,i) )\\
\beta^\tau(u,i) =& \sum_{u'} \Big( \beta^\tau(u', i+1) + Q_{i+1}' \delta(u'=\tau) \beta(u', i+1)\Big)\exp( f_\theta(u',u,X,i) ).
\end{align*}
Let $a$ and $b$ denote the labels at two adjacent sequence positions, then the gradient of equation (\ref{eq:grad}) with respect to parameter $T$ is
\begin{align*}
\frac{ \partial s(P_\theta^l, D^\tau)}{\partial T_{a,b}} = \sum_{i=1}^L \Big( \tilde{\phi}(a,b,i)\exp(f_\theta(a,b,X,i))   \Big)
\end{align*}
where 
\begin{align*}
\tilde \phi (a,b,i) = Q_i' \delta(y_i=\tau) \frac{\alpha(a, i-1) \beta(b,i)}{Z(X)} + \frac{\alpha^\tau(a, i-1) \beta(b,i)}{Z(X)} + \frac{\alpha(a, i-1) \beta^\tau(b,i)}{Z(X)} - \frac{\alpha(a, i-1) \beta(b,i)}{Z(X)}\mathcal{C} 
\end{align*}

The gradient of equation (\ref{eq:grad}) with respect to parameter $U$ is:
\begin{align*}
\frac{ \partial s(P_\theta^l, D^\tau)}{\partial U_{a,h}} = \sum_{i=1}^L \Big( \tilde{\Phi}(a,i) A_{a,h}(X,i,W)  \Big),
\end{align*}
where
\begin{align}
\tilde{\Phi}(a,i) = \frac{\alpha^\tau(a,i)\beta(a,i)}{Z(X)} + \frac{\alpha(a,i)\beta(a,i)^\tau}{Z(X)} - \frac{\alpha(a,i)\beta(a,i)}{Z(X)} \mathcal{C}
\label{eq:s23}
\end{align}

\subsection{Complexity analysis}
The gradient of the labelwise accuracy function is derived in \citet{gross2006training}. While all the three training methods have the same space complexity $O(|\Sigma|\cdot L)$, their time complexity is different. Specifically, the time complexity of calculating log-likelihood, labelwise accuracy, and the polynomial approximation of AUC is $O(|\Sigma|^2\cdot L)$, $O(|\Sigma|^2 \cdot L)$ and $O(d^2\cdot |\Sigma|^3\cdot L)$, respectively. Since DCNN is used in DeepCNF, we may not be able to solve the training problem to global optimum. Instead we use the L-BFGS~\citep{liu1989limited} algorithm to find a suboptimal solution. 

The running time of maximum-AUC training is approximately linear when the sequence length is much larger than the number of labels and the degree of the polynomial approximation. When the degree $d$ is larger, we can approximate the loss function better, but the approximation itself becomes less smooth and more challenging to optimize. A large $d$ also increases model complexity, which makes it easier to overfit. In our experiments, along with the increase of $d$, the training AUC always improves, but the test AUC drops after $d = 15$. 

\section{Results}
See Appendix for the introduction of the three protein sequence labeling problems.
This section presents our experimental results of the AUC-trained DeepCNF models on these problems. This section contains only comparison of three training methods on the three protein sequence labeling problem. See Appendix for comparison of our AUC-trained DeenCNF with the other state-of-the-art predictors of the three problems.
\subsection{Dataset}
To use a set of non-redundant protein sequences for training and test, we pick one representative sequence from each protein superfamily defined in CATH~\citep{sillitoe2015cath} or SCOP~\citep{andreeva2014scop2}. The test proteins are in different superfamilies than the training proteins, so we can reduce the bias incurred by the sequence profile similarity between the training and test proteins. The publicly available JPRED~\citep{drozdetskiy2015jpred4} dataset(http://www.compbio.dundee.ac.uk/jpred4/about.shtml) satisfies such a condition, which has 1338 training and 149 test proteins, respectively, each belonging to a different superfamily. 
We train the DeepCNF model using the JPRED training set and conduct 
7-fold cross validation to determine the model hyper-parameters for each training method. 

We also evaluate the predictive performance of our DeepCNF models on the CASP10~\citep{kryshtafovych2014assessment} and CASP11~\citep{joo2015template} test targets (merged to a single CASP dataset) and the recent CAMEO~\citep{haas2013protein} hard test targets. To remove redundancy, we filter the CASP and CAMEO datasets by removing those targets sharing $>$25\% sequence identity with the JPRED training set. This result in 126 CASP and 147 CAMEO test targets, respectively. See Appendix for their test results.

\vspace{-1.2in}
\subsection{Evaluation criteria}
\vspace{0.8in}

\begin{figure}[H]
\begin{center}
\includegraphics[width=70mm]{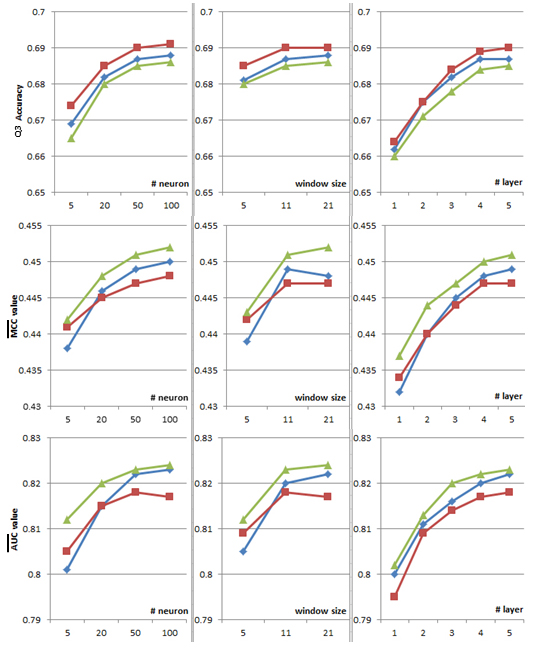}
\end{center}
\vspace{-10pt}
\caption{Q3 accuracy, mean Mcc and AUC of solvent accessibility (ACC) prediction with respect to the DCNN architecture: (left) the number of neurons, (middle) window size, and (right) the number of hidden layers. Training methods: maximum likelihood (blue), maximum labelwise accuracy (red) and maximum AUC (green).}
\label{fig:res1}
\end{figure}
\vspace{-5pt}
We use Q$x$ to measure the accuracy of sequence labeling  where $x$ is the number of different labels for a prediction task. Q$x$ is defined as the percentage of residues for which the predicted labels are correct. 
In particular, we use Q3 accuracy for ACC prediction, Q8 accuracy for SS8 prediction and Q2 accuracy for disorder prediction.

From TP (true positives), TN (true negatives), FP (false positives) and FN (false negatives), we may also calculate sensitivity (sens), specificity (spec), precision (prec) and Matthews correlation coefficient (Mcc) as $\frac{TP}{TP + FN}, \frac{TN}{TN+FP}, \frac{TP}{TP+FP}$ and $\frac{TP\times TN - FP\times FN}{\sqrt{(TP+FP)(TN+FP)(TP+FN)(TN+FN)}}$, respectively. We also use AUC as a measure.  Mcc and AUC are generally regarded as balanced measures which can be used on class-imbalanced data. Mcc ranges from −1 to +1, with +1 representing a perfect prediction, 0 random prediction and −1 total disagreement between prediction and ground truth. AUC has a minimum value 0.5 and the best value 1.0. When there are more 2 different labels in a labeling problem, we may also use mean $Mcc$ (denoted as $\bar{Mcc}$) and mean $AUC$ (denoted as $\bar{AUC}$), which are averaged over all the different labels.

\subsection{Performance}
The architecture of the DCNN in DeepCNF model is mainly determined by the following 3 factors (see Figure 1): (i) the number of hidden layers; (ii) the number of different neurons at each layer; and (iii) the window size at each layer. 

\begin{figure}[ht]
\begin{minipage}[b]{0.49\linewidth}
\includegraphics[width=\textwidth]{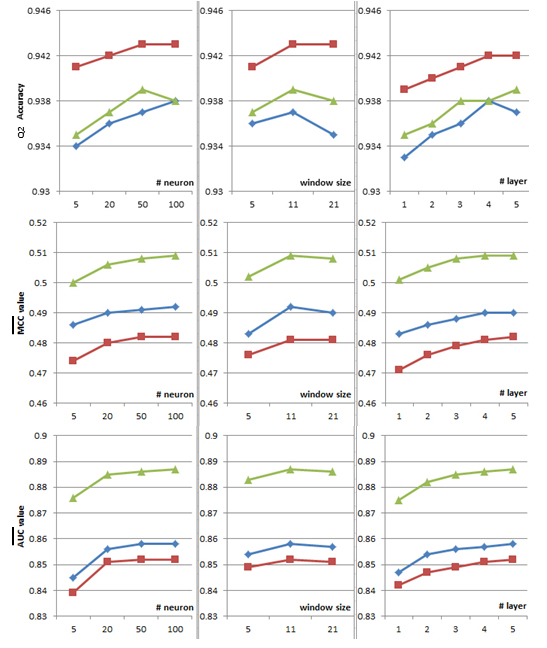}
\vspace{-10pt}
\caption{Q2 accuracy, mean Mcc and AUC of disorder (DISO) prediction with respect to the DCNN architecture: (left) the number of neurons, (middle) window size, and (right) the number of hidden layers. Training methods: maximum likelihood (blue), maximum labelwise accuracy (red) and maximum AUC (green).}
\label{fig:figure1}
\end{minipage}
\hspace{0.3cm}
\begin{minipage}[b]{0.49\linewidth}
\includegraphics[width=\textwidth]{Final_Figures/DISO_Fig.jpg}
\vspace{-10pt}
\caption{Q8 accuracy, mean Mcc and AUC of 8-state secondary structure (SS8) prediction with respect to the DCNN architecture: (left) the number of neurons, (middle) window size, and (right) the number of hidden layers. Training methods: maximum likelihood (blue), maximum labelwise accuracy (red) and maximum AUC (green).}
\label{fig:figure2}
\end{minipage}
\end{figure}


We conduct 7-fold cross-validation for each possible DCNN architecture, each training method, and each labeling problem using the JPRED dataset. To simplify the analysis, we use the same number of neurons and the same windows size for all hidden layers. By default we use 5 hidden layers, each with 50 different hidden neurons and windows size 11. 

Overall, as shown in Figures 2 to 4, when the labels are almost equally distributed, there is no big difference among the three training methods. On the other hand, when the label distribution is highly imbalanced, maximum-AUC achieves higher mean Mcc and AUC than the other two training methods (especially maximum labelwise accuracy). 
Our DeepCNF model reaches peak performance when it has 4 to 5 hidden layers, 50 to 100 different hidden neurons at each layer, and windows size 11. Further increasing the number of layers, the number of different hidden neurons, and the windows size does not result in significant improvement in Q$x$ accuracy, mean Mcc and AUC, regardless of the training method.

For ACC prediction, as shown in Figure 2, since the three labels are equally distributed, no matter what training methods are used, the best Q3 accuracy, the best mean Mcc and the best mean AUC are 0.69, 0.45, 0.82, respectively; 
For DISO prediction, since the two labels are highly imbalanced, as shown in Figure 3, although all three training methods have similar Q2 accuracy 0.94, maximum-AUC obtains mean Mcc and AUC at 0.51 and 0.89, respectively, greatly outperforming the other two;
For SS8 prediction, as shown in Figure 4, since there are three rare labels (i.e., G for 3-10 helix, B for beta-bridge, and I for pi-helix), maximum-AUC has the overall mean Mcc at 0.44 and mean AUC at 0.86, respectively, much better than maximum labelwise accuracy, which has mean Mcc at 0.41 and mean AUC less than 0.8, respectively.

\section{Conclusion}
We have presented a novel training algorithm that directly maximizes the empirical AUC to learn DeepCNF model (DCNN+CRF) from imbalanced structured data. We also studied the behavior of three training methods: maximum-likelihood, maximum labelwise accuracy, and maximum-AUC, on three real-world protein sequence labeling problems, in which the label distribution varies from equally distributed to highly imbalanced. Evaluated by AUC and Mcc, our maximum-AUC training method achieves the state-of-the-art performance in predicting solvent accessibility, disordered regions, and 8-state secondary structure. 

Instead of using a linear-chain CRF, we may model a protein by Markov Random Fields (MRF) to capture long-range residue interactions~\citep{xu2015protein}. As suggested in \citet{schlessinger2007natively}, the predicted residue-residue contact information could further contribute to disorder prediction under the MRF model. In addition to the three protein sequence labeling problems tested in this work, our maximum-AUC training algorithm could be applied to many sequence labeling problems with imbalanced label distributions~\citep{he2009learning}. For example, in post-translation modification (PTM) site prediction, the phosphorylation and methylation sites occur much less frequently than normal residues~\citep{blom2004prediction}.

\subsubsection*{Acknowledgments}

\bibliography{iclr2016_conference}
\bibliographystyle{iclr2016_conference}

\newpage
\section*{Appendix}
\setcounter{equation}{0}
{\bf S1. Three protein sequence labeling problems}\\
We employ three important protein sequence labeling problems to test our DeepCNF models trained by three different methods: solvent accessibility (ACC) prediction, disorder (DISO) prediction, and 8-state protein secondary structure (SS8) prediction. A protein sequence consists of a collection of sequentially-linked residues. We want to predict a label for each residue from the sequence information. Below we briefly introduce each problem, especially how to calculate the true label.

\emph{ACC}. We used DSSP~\citep{kabsch1983dictionary} to calculate the absolute accessible surface area for each residue in a protein and then normalize it by the maximum solvent accessibility to obtain the relative solvent accessibility (RSA)~\citep{chothia1976nature}. Solvent accessibility of one residue is classified into 3 labels: buried (B) for RSA from 0 to 10), intermediate (I) for RSA from 10 to 40 and exposed (E) for RSA from 40 to 100. The ratio of these three labels is around 1:1:1.

\emph{DISO}. Following the definition in \citet{monastyrskyy2011evaluation}, we label a residue as disordered (label 1) if it is in a segment of more than three residues missing atomic coordinates in the X-ray structure. Otherwise it is labeled as ordered (label 0). The distribution of these two labels (ordered vs. disordered) is 94:6.

\emph{SS8}. The 8-state protein secondary structure is calculated by DSSP~\citep{kabsch1983dictionary}. In particular, DSSP assigns 3 types for helix (G for 310 helix, H for alpha-helix, and I for pi-helix), 2 types for strand (E for beta-strand and B for beta-bridge), and 3 types for coil (T for beta-turn, S for high curvature loop, and L for irregular). The distribution of these 8 labels (H,E,L,T,S,G,B,I) is 35:22:19:11:8:4:1:1.

\paragraph{Existing work.}
Quite a few methods have been developed to predict ACC, DISO, and SS8~\citep{magnan2014sspro,jones2015disopred3,wang2011protein}. Many of them used networks (NN)~\citep{qian1988predicting} or support vector machines (SVM)~\citep{hirose2007poodle}. Recently, \citet{eickholt2013dndisorder} applied a deep belief network (DBN)~\citep{hinton2006fast} to DISO prediction, and \citet{zhou2014deep} reported a supervised generative stochastic network (GSN)~\citep{bengio2013deep} for SS8 prediction. Besides maximum-AUC training, our work differs from them as follows.

Our method differs from Cheng’s work on DISO prediction: (a) we use DCNN while Cheng uses DBN. DCNN is better than DBN in capturing a longer-range of sequential information; and (b) our method considers the correlation of the “ordered/disordered” states of sequentially-adjacent residues while Cheng’s method does not. 

Our method differs from Zhou’s work on SS8 prediction: (a) our method places only input features at a visible layer and treats the SS labels as hidden states while Zhou’s method places both the input features and SS labels in a visible layer; (b) our method explicitly models the SS label interdependency while Zhou’s method does not; (c) our method directly calculates the conditional probability of SS labels on input features while Zhou’s method uses sampling; and (d) our method trains the model parameter simultaneously from end to end while Zhou’s method trains the model parameters layer-by-layer.

\paragraph{Input features.}
Given a protein sequence, we use the same feature set for the prediction of ACC, DISO, and SS8. There are two types of features: residue-related feature and evolution-related feature. 

\emph{Residue-related features}. (a) amino acid identity represented as a binary vector of 20 elements; (b) amino acid physic-chemical properties (7 values from Table 1 in \citet{meiler2001generation}); propensity of being at endpoints of a secondary structure segment (11 values from Table 1 in \citet{duan2008position}; (d)correlated contact potential (40 values from Table 3 in \citet{tan2006statistical} and (e) AAindex (5 values from Table 2 in \citet{atchley2005solving}). These features may allow for a richer representation of amino acids~\citep{ma2015acconpred}.

\emph{Evolution-related features}. We use PSSM (position specific scoring matrix) generated by PSI-BLAST~\citep{blastpsi} to encode the evolutionary information of the sequence under prediction. We also use the HHM profile generated by HHpred~\citep{soding2005protein}, which is complementary to PSSM to some degree. 

{\bf S2. More details about the DeepCNF model}\\
As shown in Fig.\ref{fig:DeepCNF} in the main text, DeepCNF has three architecture hyper-parameters: (a) the number of neurons at each layer; (b) the window size at each layer; and (c) the number of hidden layers. We train the model parameters (i.e., $U, T, W$) simultaneously. We first calculate the gradient for parameter $U, T$ and then for parameter $W$. Below we explain how to calculate the DeepCNF in a
feed-forward way and the gradient by back-propagation.

\begin{figure}[h]
\begin{center}
\includegraphics[width=120mm]{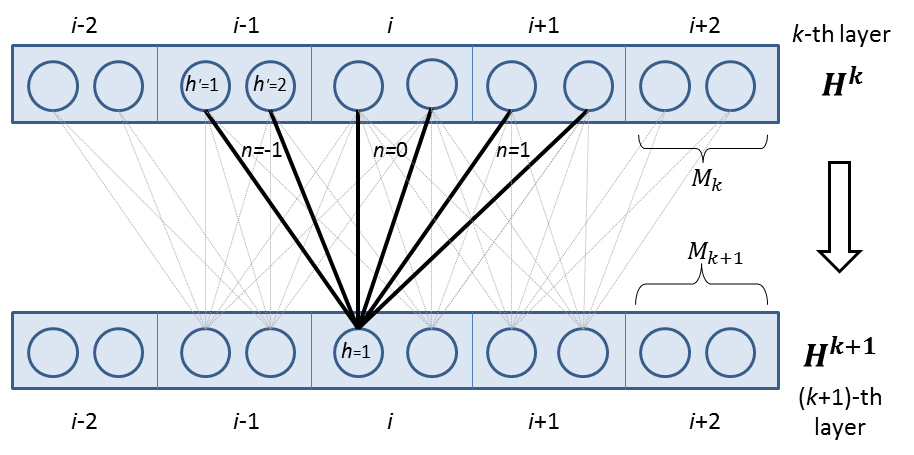}
\end{center}
\caption{The feed-forward connection between two adjacent layers of DCNN.}
\label{fig:FeedForw}
\end{figure}

{\bf S2.1 Feed-forward function of DCNN (deep convolutional neural network)}\\
Appendix Fig.~\ref{fig:FeedForw}  shows two adjacent layers of DCNN. Let $M_k$ be the number of neurons for a single position of the $k$-th layer. Let $X_i(h)$ be the $h$-th feature at the input layer for residue $i$ and $H_i^k(h)$ denote the output value of the $h$-th neuron of position $i$ at layer $k$. When $k=1, H^k$ is actually the input feature $X$. Otherwise, $H^k$ is a matrix with dimension $L\times M_k$. Let $2N_k+1$ be the window size at the $k$-th layer. Mathematically, $H_i^k(h)$ is defined as follows:
\begin{align*}
H_i^k(h) =& X_i(h)  && \text{if } k=1 \\
H_i^{k+1}(h) =& \pi\Big( \sum_{n=-N_k}^{N_k} \sum_{h'=1}^{M_k}(H_{i+n}^k(h') * W_n^k(h,h')) \Big) && \text{if } k < K \\
A_h(X, i, W) =& H_i^k(h)  && \text{if } k = K. 
\end{align*}
Meanwhile, $\pi$ is the activation function, either the sigmoid or the tanh. $W_n^k (-N_k \le n \le N_k)$ is a 2D weight matrix fir the connections between the neurons of position $i$ at layer $k$ and the neurons of position $i+1$ at layer $k+1$. $W_n^k$ is shared by all the positions in the same layer, so it is position-independent. Here $h$ and $h'$ index two neurons at the $k$-th and $(k+1)$-th layers, respectively.  

\begin{figure}[h]
\begin{center}
\includegraphics[width=120mm]{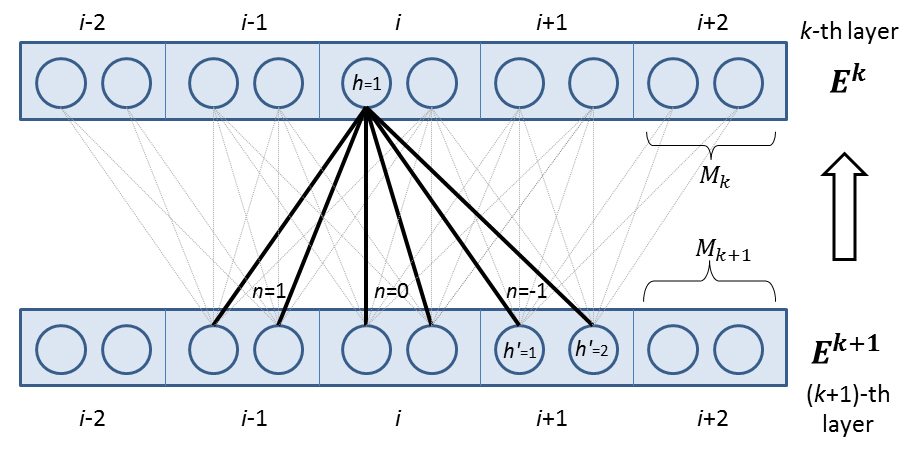}
\end{center}
\caption{Illustration of how to calculate the gradient of DCNN from layer $k + 1$ to layer $k$.}
\label{fig:Feedback}
\end{figure}

{\bf S2.2 Calculation of gradient by back-propagation}\\
The error function from the CRF part at position $i$ for a certain label $u$ is 
\[
E_i(u) = \sum_{\mu=0}^d \sum_{l=0}^\mu \mathcal{Y}_{\mu l} \Big( \tilde{\phi}_s^l(\mu,i) v(P_\theta^{\mu-l}, D^{!\tau}) + s(P_\theta^l, D^\tau) \tilde{\phi}_v^{\mu-l}(\mu, i) \Big), 
\]
where $\tilde{\phi}_s^l$ and $\tilde{\phi}_v^{u-l}$ are derived according to equation (\ref{eq:s23}) with respect to function $s(P_\theta^l, D^\tau)$ and $v(P_\theta^{\mu-l}, D^{!\tau})$, respectively.  As show in Fig.~\ref{fig:Feedback}, we can calculate the neuron error values as well as the gradients at the $k$-th layer by back-propagation as follows:
\begin{align*}
E_i^k(h) =& \eta( H_i^k(h) ) * \sum_u \Big( E_i(u) * U_{a,h} ) &&\text{if } k = K \\
E_i^k(h) =& \eta( H_i^k(h) ) * \sum_{n=-N_k}^{N_k} \sum_{h'=1}^{M_{k+1}} \Big( E_{i+n}^{k+1}(h') * W_n^k(h', h) \Big) && \text{if } k < K,
\end{align*}
where $\eta$ is the derivative of the activation function $\pi$. In particular, it is $\eta(x) = (1-x)x$ and $\eta(x) = 1-x*x$ for the sigmoid and tanh function, respectively. $E^k$ is the neuron error value matrix at the $k$-th layer, with dimension $L\times M_k$. Finally, the gradient of the parameter $W$ at the $k$-th layer is 
\begin{align*}
\nabla_{W_n^k(h, h')} = \sum_{i=1}^L \Big( E_i^{k+1}(h) * H_{i+n}^k(h') \Big)
\end{align*}

{\bf S3. Performance comparison with the state-of-the-art predictors}\\
\paragraph{Programs to compare.}
Since our method is \textit{ab initio}, we do not compare it with consensus-based or template-based methods. Instead, we compare our method with the following ab initio predictors: (i) for ACC prediction, we compare to SPINE-X~\citep{faraggi2009improving} and ACCpro5-ab~\citep{magnan2014sspro}. SPINE-X uses neural networks (NN) while ACCpro5-ab uses bidirectional recurrent neural network (RNN); (ii) for DISO prediction, we compare to DNdisorder~\citep{eickholt2013dndisorder} and DisoPred3-ab~\citep{jones2015disopred3}. DNdisorder uses deep belief network (DBN) while DisoPred3-ab uses support vector machine (SVM) and NN for prediction; (iii) for SS8 prediction, we compare our method with SSpro5-ab~\citep{magnan2014sspro} and RaptorX-SS8~\citep{wang2011protein}. SSpro5-ab is based on RNN while RaptorX-SS8 uses conditional neural field (CNF)~\citep{peng2009conditional}. We cannot evaluate Zhou’s method~\citep{zhou2014deep} since it is not publicly available.

\paragraph{Overall evaluation.}
Here we only compare our AUC-trained DeepCNF model (trained by the JPRED data) to the other state-of-the-art methods on the CASP and CAMEO datasets. 
As shown in Tables 1 to 3, our AUC-trained DeepCNF model outperforms the other predictors on all the three sequence labeling problems, in terms of the Q$x$ accuracy, Mcc and AUC. When the label distribution is highly imbalanced, our method greatly exceeds the others in terms of Mcc and AUC. Specifically, for DISO prediction on the CASP data, our method achieves 0.53 Mcc and 0.88 AUC, respectively, greatly outperforming DNdisorder (0.37 Mcc and 0.81 AUC) and DisoPred3\_ab (0.47 Mcc and 0.84 AUC). For SS8 prediction on the CAMEO data, our method obtains 0.42 Mcc and 0.83 AUC, respectively, much better than SSpro5\_ab (0.37 Mcc and 0.78 AUC) and RaptorX-SS8 (0.38 Mcc and 0.79 AUC).

\paragraph{sensitivity, specificity, and precision.}
Tables 4 and 5 list the sensitivity, specificity, and precision on each label obtained by our method and the other competing methods evaluated on the merged CASP and CAMEO data. Overall, at a high specificity level, our method obtains compatible or better precision and sensitivity for each label, especially for those rare labels such as G, I, B, S, T for SS8, and disorder state for DISO. Taking SS8 prediction as an example, for pi-helix (I), our method has sensitivity and precision 0.18 and 0.33 respectively, while the second best method obtains 0.03 and 0.12, respectively. For beta-bridge (B), our method obtains sensitivity and precision 0.13 and 0.42, respectively, while the second best method obtains 0.07 and 0.34, respectively.

\begin{table}[H]
\caption{Performance of solvent accessibility (ACC) prediction on the CASP and CAMEO data. Sens, spec, prec, Mcc and AUC are averaged on the 3 labels. The best values are shown in bold. }
\label{table:1}
\begin{center}
\begin{tabular}{*{14}{|l}|}
\cline{1-14}
\multicolumn{2}{|c|}{} & \multicolumn{6}{ |c| }{CASP} & \multicolumn{6}{ |c| }{CAMEO} \\ 
\cline{3-14}
\multicolumn{2}{|c|}{Method} & Q3 & Sens & Spec & Prec & Mcc & AUC & Q3 & Sens & Spec & Pre & Mcc & AUC \\ 
\cline{1-14}
\multicolumn{2}{ |c| }{OurMethod} &  \bf 0.69 &\bf 0.65 &\bf 0.82 &\bf 0.64 &\bf 0.47 &\bf 0.82 &\bf 0.66 &\bf 0.62 &\bf 0.81 &\bf 0.62 &\bf 0.43 &\bf 0.80 \\ 
\cline{1-14}
\multicolumn{2}{ |c| }{SPINE-X} & 0.63 &0.59 &0.80 &0.59 &0.42 &0.78 &0.61 &0.58 &0.78 &0.57 &0.39 &0.75\\
\cline{1-14}
\multicolumn{2}{ |c| }{ACCpro5\_ab} & 0.62 &0.58 &0.81 &0.57 &0.41 &0.76 &0.59 &0.55 &0.79 &0.55 &0.36 &0.73 \\
\cline{1-14}
\end{tabular}
\end{center}
\end{table}

\begin{table}[H]
\caption{Performance of order/disorder (DISO) prediction on the CASP and CAMEO data.}
\label{table:2}
\begin{center}
\begin{tabular}{*{14}{|l}|}
\cline{1-14}
\multicolumn{2}{|c|}{} & \multicolumn{6}{ |c| }{CASP} & \multicolumn{6}{ |c| }{CAMEO} \\ 
\cline{3-14}
\multicolumn{2}{|c|}{Method} & Q2 & Sens & Spec & Prec & Mcc & AUC & Q2 & Sens & Spec & Pre & Mcc & AUC \\ 
\cline{1-14}
\multicolumn{2}{ |c| }{OurMethod} &  \bf 0.94 &\bf 0.74 &\bf 0.74 &\bf 0.75 &\bf 0.53 &\bf 0.88 &\bf 0.94 &\bf 0.73 &\bf 0.73 &\bf 0.74 &\bf 0.47 &\bf 0.86 \\ 
\cline{1-14}
\multicolumn{2}{ |c| }{DisoPred3\_ab} & 0.94 &0.67 &0.67 &0.72 &0.47 &0.84 &0.94 &0.71 &0.71 &0.71 &0.42 &0.83 \\
\cline{1-14}
\multicolumn{2}{ |c| }{DNdisorder} & 0.94 &0.73 &0.73 &0.70 &0.37 &0.81 &0.94 &0.72 &0.72 &0.68 &0.36 &0.79 \\
\cline{1-14}
\end{tabular}
\end{center}
\end{table}

\begin{table}[H]
\caption{Performance of 8-state secondary structure (SS8) prediction on the CASP and CAMEO data.}
\label{table:3}
\begin{center}
\begin{tabular}{*{14}{|l}|}
\cline{1-14}
\multicolumn{2}{|c|}{} & \multicolumn{6}{ |c| }{CASP} & \multicolumn{6}{ |c| }{CAMEO} \\ 
\cline{3-14}
\multicolumn{2}{|c|}{Method} & Q8 & Sens & Spec & Prec & Mcc & AUC & Q8 & Sens & Spec & Pre & Mcc & AUC \\ 
\cline{1-14}
\multicolumn{2}{ |c| }{OurMethod} &  \bf 0.71 &\bf 0.48 &\bf 0.96 &\bf 0.56 &\bf 0.44 &\bf 0.85 &\bf 0.69 &\bf 0.45 &\bf 0.95 &\bf 0.54 &\bf 0.42 &\bf 0.83 \\ 
\cline{1-14}
\multicolumn{2}{ |c| }{RaptorX-SS8} & 0.65 &0.42 &0.95 &0.50 &0.41 &0.81 &0.64 &0.40 &0.94 &0.48 &0.38 &0.79 \\
\cline{1-14}
\multicolumn{2}{ |c| }{SSpro5\_ab} & 0.64 &0.41 &0.95 &0.48 &0.40 &0.79 &0.62 &0.38 &0.94 &0.46 &0.37 &0.78 \\
\cline{1-14}
\end{tabular}
\end{center}
\end{table}

\begin{table}[H]
\caption{Sensitivity, specificity, and precision of each solvent accessibility (ACC) label, tested on the combined CASP and CAMEO data.}
\label{table:4}
\begin{center}
\begin{tabular}{*{11}{|l}|}
\cline{1-11}
\multicolumn{2}{|c|}{ACC} & \multicolumn{3}{ |c| }{Sensitivity} & \multicolumn{3}{ |c| }{Specificity} &  \multicolumn{3}{ |c| }{Precision}\\ 
\cline{3-11}
\multicolumn{2}{|c|}{Label} & Our & SpX$^*$ & Acc5$^{**}$ & Our & SpX & Acc5 & Our & SpX & Acc5 \\
\cline{1-11}
\multicolumn{2}{ |c| }{B} &  \bf 0.77 & 0.74 & 0.75 & \bf 0.82 & 0.81 & 0.80 & \bf 0.67 & 0.63 & 0.62 \\ 
\cline{1-11}
\multicolumn{2}{ |c| }{M} & \bf 0.45 &0.36 &0.34 & \bf 0.80 &0.78 &0.79 &\bf 0.54 &0.48 &0.46 \\
\cline{1-11}
\multicolumn{2}{ |c| }{E} & \bf 0.71 &0.67 &0.63 &\bf 0.82 &0.79 &0.80 &\bf 0.67 &0.62 &0.61 \\
\cline{1-11}
\end{tabular}
\begin{tablenotes} \small  \item * SPINEX, ** ACCpro5\_ab \end{tablenotes}
\end{center}
\end{table}

\begin{table}[H]
\caption{Sensitivity, specificity, and precision of each disorder label on the combined CASP and CAMEO data.}
\label{table:5}
\begin{center}
\begin{tabular}{*{11}{|l}|}
\cline{1-11}
\multicolumn{2}{|c|}{DISO} & \multicolumn{3}{ |c| }{Sensitivity} & \multicolumn{3}{ |c| }{Specificity} &  \multicolumn{3}{ |c| }{Precision}\\ 
\cline{3-11}
\multicolumn{2}{|c|}{Label} & Our & Diso$^*$ & DN$^{**}$ & Our & Diso & DN & Our & DISO & DN \\
\cline{1-11}
\multicolumn{2}{ |c| }{0} &  \bf 0.96	 & 0.96 & 0.89 & 0.51 & 0.41 & \bf 0.55 & \bf 0.95 & 0.94 & 0.93 \\ 
\cline{1-11}
\multicolumn{2}{ |c| }{1} & 0.51 &0.41 &\bf 0.55 &\bf 0.96 &0.96 &0.89 &\bf 0.54 &0.51 &0.47 \\
\cline{1-11}
\end{tabular}
\begin{tablenotes} \small  \item * DisoPred3\_ab;  ** DNdisorder \end{tablenotes}
\end{center}
\end{table}

\begin{table}[H]
\caption{Sensitivity, specificity, and precision of each 8-state secondary structure label on the combined CASP and CAMEO data.}
\label{table:6}
\begin{center}
\begin{tabular}{*{11}{|l}|}
\cline{1-11}
\multicolumn{2}{|c|}{SS8} & \multicolumn{3}{ |c| }{Sensitivity} & \multicolumn{3}{ |c| }{Specificity} &  \multicolumn{3}{ |c| }{Precision}\\ 
\cline{3-11}
\multicolumn{2}{|c|}{Label} & Our & Rapt$^*$ & SSp5$^{**}$ & Our & Rapt & SSp5 & Our & Rapt & SSp5 \\
\cline{1-11}
\multicolumn{2}{ |c| }{H} &  \bf 0.91 &0.89 &0.90 &0.92 &\bf 0.93 &0.93 &\bf 0.85 &0.84 &0.84 \\
\cline{1-11}
\multicolumn{2}{ |c| }{G} & \bf 0.28 &0.21 &0.19 &\bf 0.99 &0.98 &0.97 &\bf 0.47 &0.43 &0.41 \\
\cline{1-11}
\multicolumn{2}{ |c| }{I} & \bf 0.18 &0.03 &0.02 &\bf 0.99 &0.98 &0.98 &\bf 0.33 &0.12 &0.06 \\
\cline{1-11}
\multicolumn{2}{ |c| }{E} & \bf 0.84 &0.78 &0.77 &\bf 0.94 &0.91 &0.89 &\bf 0.73 &0.72 &0.69 \\
\cline{1-11}
\multicolumn{2}{ |c| }{B} & \bf 0.13 & 0.05 &0.07 &\bf 0.99 &0.99 &0.99 &\bf 0.42 &0.33 &0.34 \\
\cline{1-11}
\multicolumn{2}{ |c| }{T} & \bf 0.56 &0.49 &0.51 &\bf 0.95 &0.93 &0.93 &\bf 0.56 &0.50 &0.49\\
\cline{1-11}
\multicolumn{2}{ |c| }{S} & \bf 0.29 &0.21 &0.18 &\bf 0.97 &0.96 &0.97 &\bf 0.51 &0.43 &0.45\\
\cline{1-11}
\multicolumn{2}{ |c| }{L} & 0.61 &0.62 &\bf 0.63 &0.86 &0.86 &\bf 0.87 &\bf 0.58 &0.58 &0.54 \\
\cline{1-11}
\end{tabular}
\begin{tablenotes} \small  \item * RaptorX-SS8;  ** SSpro5\_ab \end{tablenotes}
\end{center}
\end{table}

\end{document}